\documentclass[11pt,a4paper]{article}
\usepackage[hyperref]{emnlp-ijcnlp-2019}
\usepackage{times}
\usepackage{latexsym}

\usepackage{tabularx}
\usepackage{amsmath,array,graphicx,amssymb}
\usepackage{adjustbox}
\usepackage{caption}
\usepackage{subcaption}
\usepackage{multirow}
\usepackage{mathtools} 
\usepackage{graphicx} 
\usepackage{tabularx}
\usepackage{breakcites}
\usepackage{caption}
\usepackage{xcolor}
\usepackage{array}
\usepackage{makecell}
\usepackage{tabularx}
\usepackage{mathrsfs}
\usepackage{subcaption}
\usepackage{scalerel}

\usepackage{float}
\usepackage{url}

\newcommand\blfootnote[1]{%
  \begingroup
  \renewcommand\thefootnote{}\footnote{#1}%
  \addtocounter{footnote}{-1}%
  \endgroup
}

\aclfinalcopy 

\title{How to Build User Simulators to Train RL-based Dialog Systems}

\author{Weiyan Shi$^{*1}$, Kun Qian$^{*1}$, Xuewei Wang$^{2}$ and Zhou Yu$^{1}$ \\
$^1$ University of California, Davis, $^2$ Carnegie Mellon University\\
\tt \{wyshi, kunqian, joyu\}@ucdavis.edu, xueweiwa@andrew.cmu.edu 
}

\date{}

\begin{document}
\maketitle
\begin{abstract}
User simulators are essential for training reinforcement learning (RL) based dialog models. The performance of the simulator directly impacts the RL policy. However, building a good user simulator that models real user behaviors is challenging. We propose a method of standardizing user simulator building that can be used by the community to compare dialog system quality using the same set of user simulators fairly.
We present implementations of six user simulators trained with different dialog planning and generation methods. We then calculate a set of automatic metrics to evaluate the quality of these simulators both directly and indirectly. We also ask human users to assess the simulators directly and indirectly by rating the simulated dialogs and interacting with the trained systems. This paper presents a comprehensive evaluation framework for user simulator study and provides a better understanding of the pros and cons of different user simulators, as well as their impacts on the trained systems. 
\blfootnote{* Equal contribution.}
\footnote{The code and data are released at \url{https://github.com/wyshi/user-simulator}.}
\end{abstract}

\section{Introduction}
Reinforcement Learning has gained more and more attention in dialog system training because it treats the dialog planning as a sequential decision problem and focuses on long-term rewards \cite{su2017sample}. However, RL requires interaction with the environment, and obtaining real human users to interact with the system is  both time-consuming and labor-intensive. Therefore, building user simulators to interact with the system before deployment to real users becomes an economical choice \cite{williams2017hybrid, li2016user}. But the performance of the user simulator has a direct impact on the trained RL policy.  Such an intertwined relation between user simulator and dialog system makes the whole process a ``chicken and egg'' problem. This naturally leads to the question of how different user simulators impact the system performance, and how to build appropriate user simulators for different tasks. 

In previous RL-based dialog system literature, people reported their system's performance, such as success rate, on their specific user simulators \cite{liu2017iterative, shi2018sentiment}, but 
the details of the user simulators  are not sufficient to reproduce the results.
 User simulators' quality can vary in multiple aspects, which could lead to unfair comparison between different trained systems.  For instance, RL systems built with more complicated user simulators will have lower scores on the automatic metrics, compared to those built using simpler user simulators. However, the good performance may not necessarily transfer when the system is tested by real users. In fact, models that have a low score but are trained on better simulators may actually perform better in real situations because they have experienced more complex scenarios. In order to obtain a fairer comparison between systems, we propose a set of standardized user simulators. We pick the popular  restaurant search task from Multiwoz \cite{budzianowski2018multiwoz} and analyze the pros and cons of different user simulator building methods.      

The potential gap between automatic metrics and real human evaluation also makes user simulator hard to build. The ideal evaluator of a dialog system should be its end-users. But as stated before, to obtain real user evaluation is time-consuming. Therefore, 
many automatic metrics have been studied to evaluate a user simulator \cite{pietquin2013survey, kobsa1994user} from different perspectives. However, we do not know how these automatic metrics correlate with human satisfaction. In this paper, we ask human users to both rate the dialogs generated by the user simulators, and interact with the dialog systems trained with them, in order to quantify the gap between the automatic metrics and human evaluation.

This paper presents three contributions: first, we annotate the user dialog acts in the restaurant domain in Multiwoz 2.0; second, we build multiple user simulators in the standard restaurant search domain and publish the code to facilitate further development of RL-based dialog system training algorithms; third, we perform comprehensive evaluations on the user simulators and trained RL systems, including automatic evaluation, human  rating simulated dialogs, human interacting with trained systems and cross study between simulators and systems, to measure the gap between automatic dialog completion metrics with real human satisfaction, and provide meaningful insights on how to develop better user simulators. 

\section{Related work}
One line of prior user simulator research focuses on agenda-based user simulator (ABUS)
\cite{schatzmann2006survey, schatzmann2007agenda, schatzmann2009hidden, li2016user} and it is most commonly used in task-oriented dialog systems. An agenda-based user simulator is built on hand-crafted rules according to an agenda provided at the beginning of a dialog. 
This mechanism of ABUS makes it easier to explicitly integrate context and agenda into the dialog planning. \citet{schatzmann2009hidden} presented a statistical hidden agenda user simulator, tested it against real users and showed that a superior result in automatic metrics does not guarantee a better result in the real situation. \citet{li2016user} proposed an agenda-based user simulator in the movie domain and published a generic user simulator building framework. In this work, we build a similar agenda-based user simulator in the restaurant domain, and focus more on analyzing the effects of using different user simulators. 


However, it's not feasible to build agenda-based user simulators for more complex tasks without an explicit agenda. Therefore, people have also studied how to build user simulators in a data-driven fashion.  \citet{he2018decoupling} fit a supervised-learning-based user simulator to perform RL training on a negotiation task. \citet{asri2016sequence} developed a seq2seq model for user simulation in the restaurant search domain, which took the dialog context into consideration without the help of external data structure. \citet{kreyssig2018neural} introduced the Neural User Simulator (NUS) which learned user behaviour from a corpus and generates natural language directly instead of semantic output such as dialog acts. However, unlike in ABUS, how to infuse the agenda into the dialog planning and assure consistency in data-driven user simulators has been an enduring challenge. 
In this paper, we present a supervised-learning-based user simulator and integrate the agenda into the policy learning. Furthermore, we compare such a data-driven method with its agenda-based counterpart. 

Another line of user simulator work treats the user simulator itself as a dialog system, and train the simulator together with the RL system iteratively \cite{liu2017iterative, shah2018building}. \citet{shah2018building} proposed the Machines Talking To Machines
(M2M) framework to bootstrap both user and system agents with dialog self-play. 
\citet{liu2017iterative} presented a method for iterative dialog policy training and address the problem of building reliable simulators by optimizing the system and the user jointly. But such iterative approach requires extra effort in setting up RL and designing reward for the user simulator, which may result in the two agents exploiting the task, and leads to  numerical instability.

Another challenging research question is how user simulator performance can be evaluated \cite{schatztnann2005effects, ai2011assessing, ai2011comparing, engelbrecht2009analysis, hashimoto2019unifying}. 
\citet{pietquin2013survey} conducted a comprehensive survey over metrics that have been used to assess  user simulators, such as perplexity and BLEU score \cite{papineni2002bleu}. However, some of the metrics are designed specifically for language generation evaluation, and as \citet{liu2016not} pointed out, these automatic metrics barely correlate with human evaluation. Therefore, \citet{ai2011assessing} involved human judges to directly rate the simulated dialog. \citet{schatzmann2009hidden} asked humans to interact with the trained systems to perform indirect human evaluation. \citet{schatztnann2005effects} proposed \textit{cross-model evaluation} to compare user simulators since human involvement is expensive. We combine the existing evaluation methods and conduct comprehensive assessments to measure the gap between automatic metrics and human satisfaction. 





\section{Dataset}
\label{sec:dataset}
We choose the restaurant domain in Multiwoz 2.0 \cite{budzianowski2018multiwoz} as our dataset, because it's the most classic domain in task-oriented dialog systems.  The system's task is to help users find restaurants, provide restaurant information and make reservations. There are a total of 1,310 dialogs annotated with  \textit{informable} slots (e.g.~\textit{food, area}) that narrow downs the restaurant choice, and \textit{requestable} slots (e.g.~\textit{address, phone}) that track users' detailed requests about the restaurant.  But because the original task in Multiwoz was to model the system response, it only contains dialog act annotation on the system-side but not on the user-side. To build user simulators, we need to model user behaviors, and therefore, we annotate the user intent in Multiwoz. In order to build user simulators, we need to model  user behavior and therefore, we  annotate the user-side dialog act in the restaurant domain of Multiwoz. 
Two human expert annotators analyze the data and agree on a set of seven user dialog acts ($UserActs$): \textit{``inform restaurant type''}, \textit{``inform restaurant type change''}, \textit{``anything else''}, \textit{``request restaurant info''}, \textit{``make reservation''}, \textit{``make reservation change time''}, and \textit{``goodbye''}.  Because the data is relatively clean and constrained in domain, the annotation is performed by designing  regular expression first and cleaned by human annotators later. We manually checked 10\% of the data (around 500 utterances) and the accuracy for automatic annotations is 94\%. These annotated user dialog acts will serve as the foundation of the user simulator action space $UserActs$. The annotated data is released to facilitate user simulator study. 

\section{User Simulator Design}
According to \citet{li2016user}, user simulator building eventually boils down to two important tasks: building  
1) a dialog manager (DM) \cite{henderson2014third, cuayahuitl2015strategic, young2013pomdp} that governs the simulator's next move; and 2) a natural language generation module (NLG) \cite{tran2017semantic, duvsek2016sequence} that translates the semantic output from dialog manager into natural language. The user simulator can adopt either \textit{agenda-based} approach or \textit{model-based} approach for the dialog manager. While for NLG, the user simulator can use the  dialog act to select pre-defined templates, retrieve user utterances from previously collected dialogs, or generate the surface form utterance directly with pre-trained language model \cite{jung2009data}.

The dialog manager module ensures the intrinsic logical consistency of the user simulator, while the NLG module controls the extrinsic language fluency. 
DM and NLG play an equally important role in the user simulator design and must go hand-in-hand to imitate user behaviours. Therefore, we propose to test different combinations of DM and NLG methods to answer the question of how to build the best user simulator. 


In task-oriented dialog systems, the user simulator's task is to complete a pre-defined goal by interacting with the system.  Multiwoz provides detailed goals for each dialog, which serves as the goal database. These goals consist of sub-tasks, such as request information or make reservation. An example goal is, ``\textit{You're looking for an Italian restaurant in the moderate price range in the east. Once you find the restaurant, you want to book a table for 5 people at 12:15 on Monday. Make sure you get the reference number.}'' 
 During initial RL experiments, we find that similar to supervised learning, the data imbalance in goals will impact the reinforce learning in the simulated tasked-oriented dialog setting. We find that 2/3 of the goals contain the sub-task ``ask info'' and the rest 1/3 are about ``make reservation''. Because the user simulators are all goal-driven, the RL policy is only able to experience the ``reservation'' scenario 1/3 of the time on average, which will result in the model favoring the ``ask info'' scenario more, especially in the early training stage. This further misleads the policy \cite{su2017sample}. Therefore, we augment the goal set with more  ``make reservation'' sub-task  from MultiWoz to make the sub-tasks of ``make reservation'' and ``ask info'' even. This augmented goal set with more even distribution serves as our goal database. We randomly sample a goal from the goal database during training. A user goal defines the agenda the user simulator needs to follow, so we'll use ``goal'' and ``agenda'' interchangeably in this paper.

\subsection{Dialog Manager}
\textbf{Agenda-based} We employ the traditional agenda-based stack-like user simulator~\cite{schatzmann2009hidden, li2016user}, where the dialog manager chooses a dialog act  among the user dialog act set $UserActs$ mentioned in Section~\ref{sec:dataset}. The dialog act transition is governed by hand-set rules and probabilities based on the initial goal. For example, after the system makes a recommendation, the user can go on to the next sub-task, or ask if there is another option. Fig.~\ref{fig:ddvrnn, restaurant} shows a typical agenda. Because the restaurant task is a user-initiated task, agenda-based simulator's first dialog act is always ``inform restaurant type''. 
\begin{figure}[htp!]
\centering
    \includegraphics[width=0.6\columnwidth, height=4.5cm]{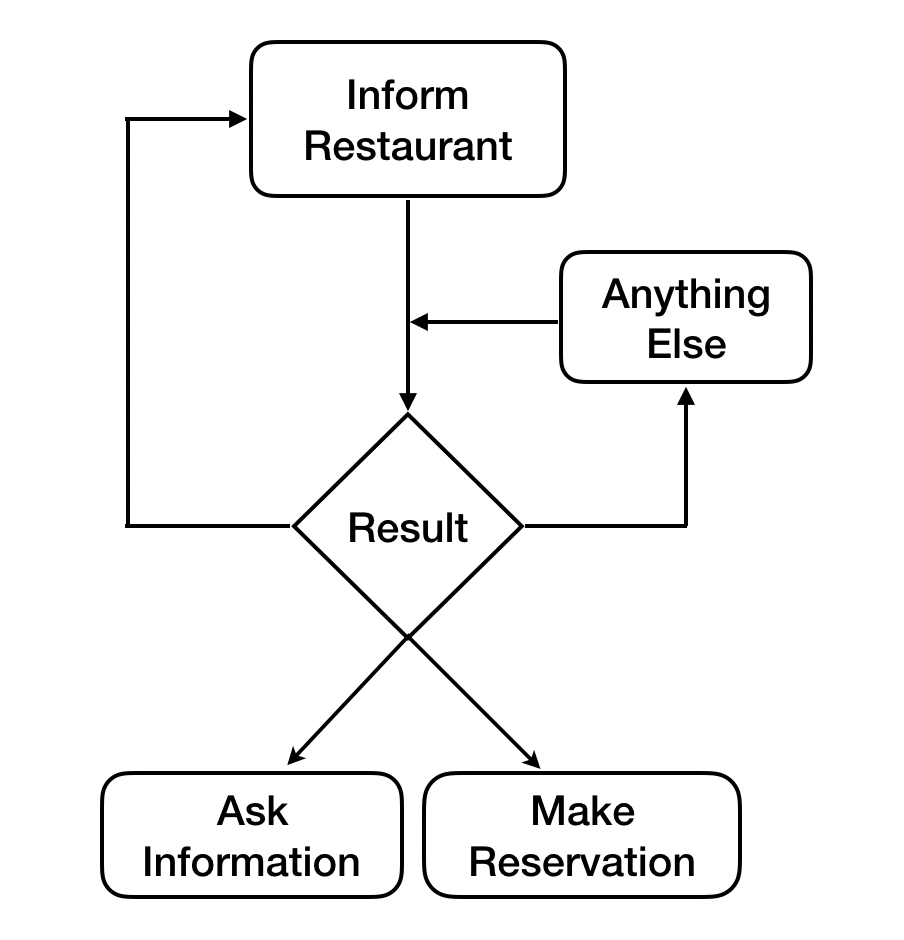}
    \caption{An example user agenda for the restaurant task}
    \label{fig:ddvrnn, restaurant}
\end{figure}
The dialog history is managed with the user dialog state by pushing and popping important slots to ensure consistency. 
Although the restaurant search task is simple, designing an agenda-based system for the task is non-trivial, because there are many corner cases to be handled. However, the advantage of building agenda-based system is that it does not require thousands of annotated dialogs.

\noindent\textbf{Model-based} It requires specific human expertise to design rules for agenda-based user simulators (compared to more easily accessible annotation), and the process is both labor-intensive and error-prone. Moreover, for complicated tasks such as negotiation, it is not practical to design rules in the policy \cite{he2018decoupling}. Therefore, we explore the possibility of building dialog manager 
with supervised learning methods. 
Compared to agenda-based simulators which require special expert knowledge, supervised learning methods require less expert involvement.
\begin{figure}[htp!]
\centering
  \includegraphics[width=0.9\columnwidth]{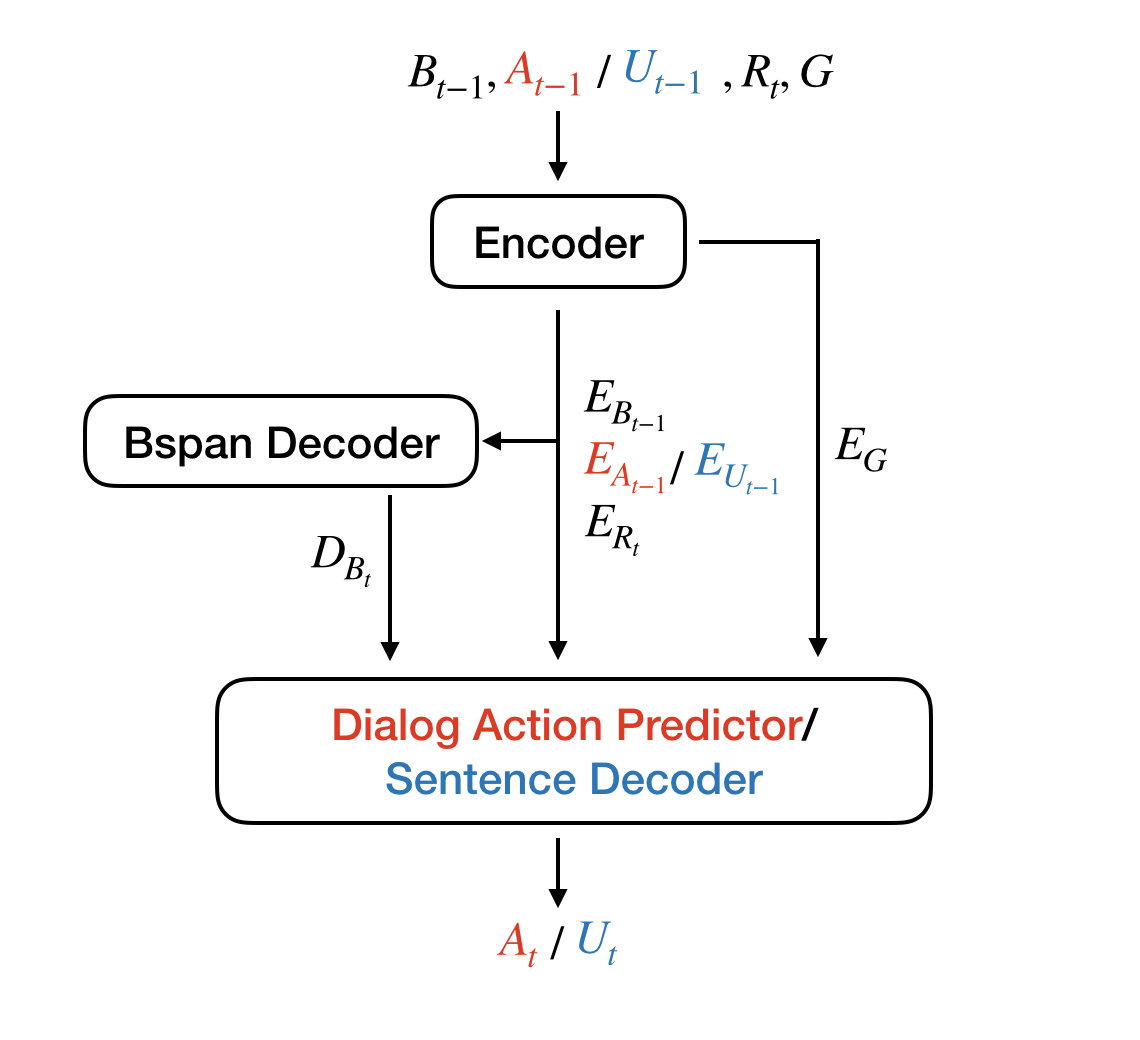}
  \caption{The end-to-end simulator and user dialog act predictor share the most part of their model, colored in black, except the decoder. All the parameters colored in red are related to the dialog act predictor and the parameters in blue color are for sentence decoder.}
  \label{fig:seq_simulator}
\end{figure}
We utilize Sequicity \cite{lei2018sequicity} to construct model-based user simulator. Sequicity is a simple seq2seq dialog system model with copy and attention mechanism. It also used belief span to track the dialog states. For example, \textit{inform:\{Name:``Caffee Uno"; Phone:``01223448620"\}} records the information that the system offers and this would be kept in belief span throughout the dialog, while \textit{request:\{``food", ``price range"\}} means the system is  asking for more information from user to locate a restaurant, which would be removed from belief span once the request is fulfilled. There are 13 types of system dialog acts. To focus on the valuable information to fit the model, we combine these dialog acts into 5 categories: \textit{\{``inform",``request",``book inform",``select", ``recommend"\}}. Similarly, we define three types of user goals \textit{``inform",``request"} and \textit{``book"}, and record them in belief span, denoted as $G$. So, at time $t$, we first update belief span with a seq2seq model, based on current system response $R_t$, previous belief state $B_{t-1}$ and previous user utterance $U_{t-1}$:
$$B_t=\textrm{seq2seq}(B_{t-1},U_{t-1},R_t)$$
Then we incorporate user goal and the context above to generate current user utterance:
$$U_t=\textrm{seq2seq}(B_{t-1},U_{t-1},R_t|B_t,G)$$
As illustrated in Fig.~\ref{fig:seq_simulator}, we build a GRU-based encoder for all the $B_{t-1}$, $U_{t-1}$, $R_t$ and the goal $G$. Then we decode the current belief span and user utterance separately. Both decoders are a one-layer GRU with copy and attention mechanism. To evaluate the dialog manager alone, we also modify the Sequicity's second decoder to generate system dialog act ($A_t$) instead of system utterances. 
$$A_t=\textrm{seq2seq}(B_{t-1},U_{t-1},R_t|B_t,G)$$

\subsection{Natural Language Generation}
Dialog act-based NLG is formalized as $U_t = M(A_t)$, where $A_t$ is the selected dialog act by the dialog manager, $U_t$ is the generated user utterance. We describe three different  dialog-act-based NLG methods. 

\noindent\textbf{Template} Template method requires human experts to write a variety of delexicalized templates for each dialog act. By searching in the templates, it translates $A_t$ into human-readable utterances. The quality of the templates have direct impact on the NLG quality.

\noindent\textbf{Retrieval} Template method suffers from limited vocabulary size and language diversity. An alternative method is Retrieval-based NLG~\cite{wu2016sequential, hu2014convolutional}. The model retrieves user utterances with $A_t$ as their dialog act in the training dataset. Following \citet{he2018decoupling},  we represent the context by a TF-IDF weighted bag-of-words vector and compute the similarity score between the candidate's context vector and the current context vector to retrieval $U_t$. 

\noindent\textbf{Generation} Generation method~\cite{wen2015stochastic, wen2015semantically} does not need expert involvement to rewrite templates, but requires  dialog act annotation similar to retrieval method. We build a vanilla seq2seq~\cite{sutskever2014sequence} model using the annotated data adding $A_t$ in the input.

\begin{table*}[htb!]

 \begin{adjustbox}{width=\textwidth}
\begin{tabular}{c|c|c|c|ccc|ccccc}
\hline
\textbf{Simulators} & \textbf{NLU} & \textbf{DM} & \textbf{NLG} & \textbf{PPL} &  \textbf{Vocab} & \textbf{Utt} &  \textbf{Hu.Fl} & \textbf{Hu.Co} & \textbf{Hu.Go} & \textbf{Hu.Div} & \textbf{Hu.All}\\ \hline \hline
Agenda-Template (AgenT)       & SL           & Agenda        & Template     &   10.32         &                  180             &    9.65     &      4.07      &     4.56     &  4.88        &    2.4   &    4.50               \\ \hline
Agenda-Retrieval (AgenR)      & SL           & Agenda        & Retrieval    &    33.90          &                  \textbf{383}             &   \textbf{11.61}        &                       3.50         &      4.22           &        4.58       &     3.9   &  3.74              \\ \hline
Agenda-Generation (AgenG)     & SL           & Agenda        & Generation   &    \textbf{7.49}          &                   159            &    8.07       &      3.32          &      3.92           &        4.64       &     2.5  &  3.36                 \\ \hline
SL-Template (SLT)         & \multicolumn{2}{c|}{SL}    & Template     &    9.32          &                   192            &   9.83      &       \textbf{4.80}         &      \textbf{4.80}           &       \textbf{4.98}        &     2.6   & \textbf{4.74}                \\ \hline
SL-Retrieval (SLR)        & \multicolumn{2}{c|}{SL}    & Retrieval    &   29.36           &                  346             &    11.06       &      4.40          &    3.99             &      4.88         &     \textbf{4.3}   & 4.01                \\ \hline
SL-End2End (SLE)       & \multicolumn{3}{c|}{End-to-End}   &   13.47           &                  205             &   10.95         &     3.32           &     2.62            &      3.18         &     2.7  &  2.64                  \\ \hline
\end{tabular}
\end{adjustbox}
\caption{Automatic metrics and human evaluation scores of different user simulators. Automatic metrics include, perplexity per word (PPL), vocabulary size (Vocab), average utterance length (Utt). Human evaluation metrics include, sentence fluency (Hu.Fl), coherent (Hu.Co), goal adherence (Hu.Go), language diversity (Hu.Div) and overall score (Hu.All).}
\label{table:simulator_results}

\end{table*}

\section{Dialog System Training Setting}
Traditionally, hand-crafted dialog acts plus slot values are used as the discrete action space in RL training \cite{raux2005let}. Dialog action space can also be on the word-level. However, previous study shows degenerate behavior when using word-level action space \cite{ zhao2019rethinking}, as it is difficult to design a reward. We choose the first approach and use the discrete action space with six system dialog acts: ``ask restaurant type'', ``present restaurant search result'', ``provide restaurant info'', ``ask reservation info'', ``inform reservation result'', ``goodbye''. Simple action masks are applied to avoid impossible actions such as making reservation before presenting a restaurant.

We use a 2-layer bidirectional-GRU with 200 hidden units to train a NLU module. For simplicity, we use the template-based method in the system's NLG module. We used policy gradient method to train dialog systems \cite{williams1992simple}. During RL training, a discounted factor of 0.9 is applied to all the experiences with the maximum number of turns to be 10. We also apply the $\epsilon$-greedy exploration strategy \cite{tokic2010adaptive}. All the RL systems use the same RL state representation, which consists of traditional dialog state  and word count vector of the current utterance.  The same reward function is used, which is $+1$ for task success, $-1$ for task failure and $-0.1$ for each additional turn to encourage the RL policy to finish the task faster rather than slower. We fix the RL model every 1,000 episodes and test for 100 dialogs to calculate the average success rate, shown in Fig.~\ref{fig:act converge curve}.

Besides RL systems, we also build a rule-based system \textit{Rule-System}, which serves as the third-party system to interact with each user simulator and generate simulated dialogs for human evaluation.  The only difference between \textit{Rule-System} and the RL-based systems is their policy selection module, where \textit{Rule-System} uses hand-crafted rules while RL-based systems use RL policy.

\section{User Simulator Evaluation}
Evaluating the quality of a user simulator is an enduring challenge. Traditionally, we report \textit{direct} automatic metrics of the user simulator, such as perplexity \cite{ai2011comparing, pietquin2013survey}. Besides, the performance of the RL system trained with a specific simulator gives us an \textit{indirect} assessment of the user simulator's ability to imitate user behaviours. 

The ultimate goal of the user simulator is to build a task-oriented RL system to serve real users. Therefore, the most ideal evaluation should be conducted by human. Therefore, we first asked human to read the simulated dialogs and rate the user simulator's performance \textit{directly}. We then hired  Amazon Mechanic Turkers (AMT) to interact with the RL systems trained with different simulators and rate their performance. 
Besides, we also performed cross study between user simulators and systems trained with different simulators to see if the systems' performance can be transferred to a different simulated setting. 
Finally, we measure the gap between the automatic metrics and human evaluation scores, and share insights on how to evaluate user simulator effectively.

\subsection{Automatic Evaluation}

\noindent\textbf{Perplexity (Direct)} Perplexity measures the language generation quality of the user simulator. The results are shown in Table~\ref{table:simulator_results}. For each simulator model, we generate 200 dialogs with the third-party \textit{Rule-System} and train a trigram language model with the data. Then we test the model and compute the perplexity with 5000 user utterances sampled from MultiWoz. Although the perplexity for retrieval models is the highest in both agenda-based and SL-based simulators, it also possesses the biggest vocabulary set and the longest average utterance length. Another common automatic metrics used to assess the language model is BLEU, but since this is a user simulator study and we don't have ground truth, BLEU score is not available.

\noindent\textbf{Vocabulary Size (Direct)} Vocabulary size is a simple and straightforward metric that measures the language diversity. As expected, 
retrieval-based models  have the biggest vocabulary set. However, Agenda-Generation has the smallest vocabulary set. The possible reason behind is that we adopt a vanilla greedy seq2seq that suffers from generating the most frequent words. SL-End2End in Table~\ref{table:simulator_results} trains the NLU, DM and NLG jointly, and therefore, the vocabulary size is slightly larger than the template-based methods.

\noindent\textbf{Average utterance length (Direct)} Average utterance length is another simple metric to assess the language model and  language diversity. As expected, retrieval-based methods are doing the best, but SL-End2End is also doing a good job in generating long sentences. 

\begin{figure}
\centering
  \includegraphics[width=0.98\columnwidth]{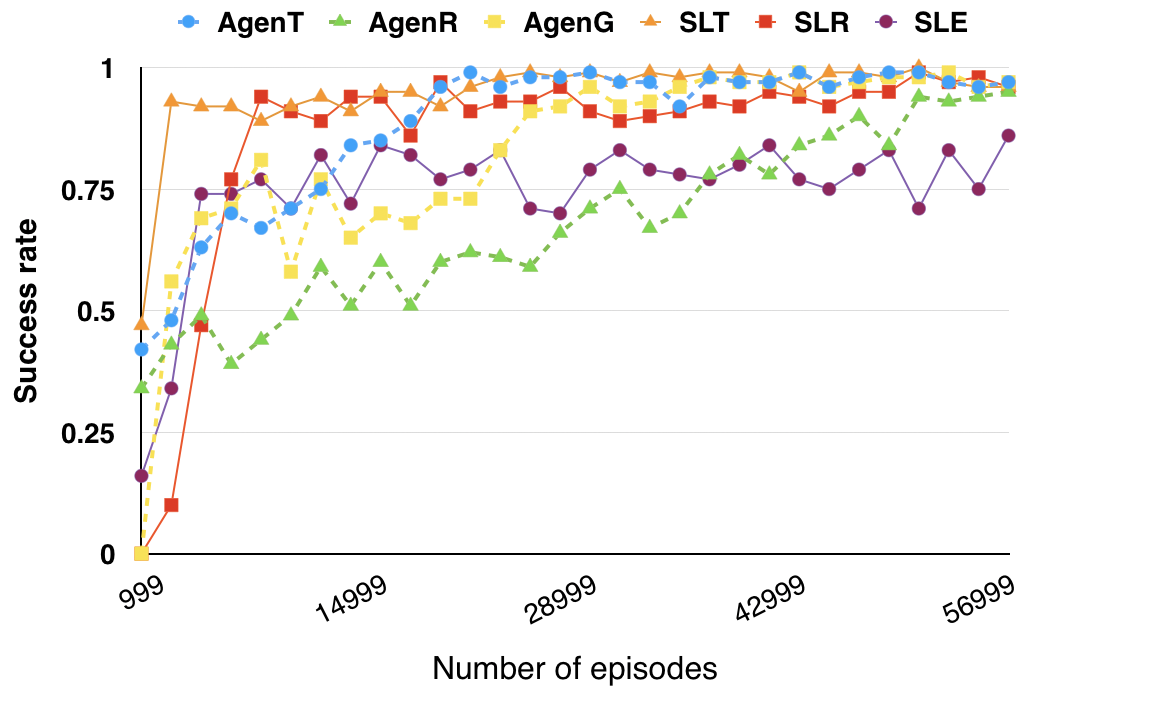}
  \vspace*{-4mm}
  \caption{Average success rate during RL training.}
  \label{fig:act converge curve}
\end{figure}

\noindent\textbf{Success Rate (Indirect)} The success rate is the most commonly used metric in reporting RL dialog system performance. Also, it can reflect the user simulator's certain behaviour. The success rate of various user simulators are shown in Fig.~\ref{fig:act converge curve}. SL-based user simulators converge faster than rule-based simulators. It can be explained by the observation that SL tries to capture the major paths in the original data, and counts those as success, instead of exploring all the possible paths like in the agenda-based simulators. In general, retrieval-based simulators converge slower than other NLG methods because retrieval-based approach has a bigger vocabulary size.

\subsection{Human Direct Evaluation}
The direct evaluation of the user simulator is conducted by asking 10 volunteers  to read the simulated dialogs between different simulators and the third-party \textit{Rule-System}. Each of the 10 volunteers would rate five randomly-selected dialogs generated from each model, and the average of the total 50 ratings is reported as the final human-evaluation score. The \textit{Rule-System} is built solely based on hand-crafted rules with no knowledge about any of the simulators, and therefore is fair to all of them. 
We design four metrics to assess the user simulator's behaviour from multiple aspects. The results are shown in Table~\ref{table:simulator_results}.

\noindent\textbf{Fluency} focuses on the language quality, such as grammar, within each utterance unit. Agenda-Template (AgenT) and SL-Template (SLT) received the two highest fluency scores because the templates are all written by human.

\begin{table*}[htb!]
\centering
\small
 \begin{adjustbox}{width=0.95\textwidth}
\begin{tabular}{l|c|c|c|c|c|c|c}
\hline
\textbf{RL System} &  \textbf{Solved Ratio} & \textbf{Satisfaction} & \textbf{Efficiency} & \textbf{Naturalness} & \textbf{Rule-likeness} &
\textbf{Dialog Length} & \textbf{Auto Success}   \\ \hline \hline
Sys-AgenT        &   0.814  $\pm 0.06 $            &      4.29  $\pm 0.20 $          &   4.35  $\pm 0.21 $             &    3.96  $\pm 0.23  $         &    4.49  $\pm 0.15  $ & 8.95 $\pm 0.38$ & \textbf{0.983 $\pm 0.01 $} \\ \hline
Sys-AgenR       &     \textbf{0.906  $\pm 0.05  $       }  &  \textbf{4.52}  $\pm 0.15  $               &    4.45 $\pm 0.16  $  &    4.23 $\pm 0.19  $            &  4.59 $\pm 0.14  $ &\textbf{8.73} $\pm 0.31$   &0.925 $\pm 0.02$ \\ \hline
Sys-AgenG      &    0.904    $\pm 0.05  $     &   4.38 $\pm 0.18  $            &      
 \textbf{4.46} $\pm 0.19  $  &    \textbf{4.33} $\pm 0.17  $ &     4.51 $\pm 0.16  $ & 9.48 $ \pm 0.45$  &0.980 $ \pm 0.01$ \\ \hline
Sys-SLT           &    0.781 $\pm 0.07  $         &     3.87 $\pm 0.22  $          &   3.81 $\pm 0.22  $           &   3.63 $\pm 0.22  $        &   \textbf{4.08}                      $\pm 0.21  $ & 9.61 $ \pm 0.76$     &0.978 $\pm 0.01$      \\ \hline
Sys-SLR        &   0.823 $\pm 0.05  $          &     4.23 $\pm 0.20  $          &  4.20 $\pm 0.10  $           &   3.99 $\pm 0.20  $          &   4.42 $\pm 0.17  $ & 8.92 $ \pm 0.70$ &0.965 $\pm 0.01$\\ 
\hline
Sys-SLE       &  0.607 $  \pm 0.06  $           &     3.42 $\pm 0.22  $          &   3.41 $\pm 0.23  $           &   3.59 $\pm 0.20  $           &    4.22 $\pm 0.20  $ & 9.44 $ \pm 0.69$  &0.798 $\pm 0.03$        \\ \hline

\end{tabular}
\end{adjustbox}
\caption{Human evaluation of  RL  systems trained with different simulators on AMT with 95\% confidence intervals. Each row represents one RL system, e.g. Sys-AgenT means the RL system trained with the AgenT simulator.}
\label{table: human evaluation on AMT}
\end{table*}

\noindent\textbf{Coherence}  focuses on the relation quality between different turns within one dialog. SL-Template (SLT) simulator performs the best in coherence, but agenda-based simulators in general are a bit more coherence than SL-based ones.

\noindent\textbf{Goal Adherence} focuses on the relation between the goal and the simulator-generated utterances. 
Both agenda-based and SL-based simulators in general stick to the goal with the exception of SL-End2End (SLE). This may be because SLE is training all the modules together and thus has more difficulty infusing the goal.

\noindent\textbf{Diversity} focuses on the language diversity between simulated dialogs of the same simulator. Each simulator will be given one diversity score. Retrieval-based methods surpass other methods in diversity, but it is not as good in fluency, while template-based methods outperform in fluency but suffer on diversity as expected. Generative methods suffer from generating generic sentences as mentioned before.

\noindent\textbf{Overall} We ask the human to rate the overall simulator quality. Except for SL-End2End, SL-based methods are favoured by human over agenda-based methods. Agenda-Template is comparable to SL-based simulators because of its fluent responses and carefully-designed policy.

\subsection{Human Indirect Evaluation on AMT}
The ultimate goal of user simulator building is to train better system policies. 
Automatic metrics such as success rate can give us a sense on the system's performance, but the ultimate evaluation should be conducted on human so that we can know the real performance of the system policy when deployed.

Motivated by this, we tested the RL systems trained with various user simulators on Amazon Mechanical Turk (AMT) \cite{miller2017parlai}, and asked Turkers to interact with the system and obtained their opinions. Each system is tested on 100 Turkers. The results are shown in Table~\ref{table: human evaluation on AMT}. The AMT interface is in the Appendix.

We also listed two common  automatic metrics in Table~\ref{table: human evaluation on AMT} to compare. The ``Dialog Length'' column shows the average dialog length of the Turker-Machine dialogs, which reflects the system's efficiency to some extent.  The ``Auto Success'' column represents the automatic success rate. It's the convergent success rate from Fig.~\ref{fig:act converge curve}, measured by freezing the policy and testing against the user simulator for 100 episodes. Previous approaches have utilized these two automatic metrics to evaluate the system's efficiency and success \cite{williams2017hybrid, shi2019unsupervised}, but we find that due to user individual difference, such automatic metrics have relatively big variances and don't always correlate with efficiency perceived by human. For example, some users tend to provide all slots in one turn, while others provide slots only when necessary; some users would even go off-the-script and ask about restaurants not mentioned in the goal. Therefore, we should caution against relying solely on the automatic metrics to represent user opinion and the best way is  to ask the users directly for their thoughts on the system's performance from multiple aspects as follows.


\begin{table*}[htb!]
    \centering
    \begin{adjustbox}{width=0.8\textwidth}
    \begin{tabular}{c||c|c|c|c|c|c}
    \hline
      Usr\textbackslash Sys   & \textbf{Sys-AgenT} & \textbf{Sys-AgenR} & \textbf{Sys-AgenG} & \textbf{Sys-SLT} & \textbf{Sys-SLR} & \textbf{Sys-SLE}  \\
         \hline
         \hline
    AgenT    & 0.975 & 0.960 & 0.790 & 0.305 & 0.300 & 0.200\\
    \hline
    AgenR & 0.540 & 0.900 & 0.785 & 0.230 & 0.230 & 0.235\\
    \hline
    AgenG & 0.725 & 0.975 & 0.950 & 0.355 & 0.300 & 0.20\\
    \hline
    SLT & {0.985} & {0.985} & {0.985} & 0.990 & 0.965 & 0.730\\
    \hline
    SLR & 0.925 & 0.975 & 0.965 & 0.975 & 0.935 & 0.630\\
    \hline
    SLE & 0.770 & 0.820 & 0.815 & 0.840 & 0.705 & 0.770\\
    \hline
    \hline
    Average & 0.820 & \textbf{0.935} & 0.882 & 0.616 & 0.573 & 0.461\\
    \hline
    \end{tabular}
    \end{adjustbox}
    \caption{Cross study results. Each row represents one user simulator, each column represents one RL system trained with a specific simulator. Each entry shows the average success rate obtained by having the user simulator interacting with the RL system for 200 times. 
    }
    \label{tab:cross_test}
\end{table*}

\noindent\textbf{Solved Ratio.} Each Turker is given a goal at the beginning, the same as in the simulated setting. At the end of the dialog, we ask the Turker if the system has solved his/her problem. There are three types of answers to this question, ``Yes'' is coded as 1, ``Partially solved'' is coded as 0.5, and ``No'' is coded as 0. Sys-AgenR is the system trained with the Agenda-Retrieval (AgenR) simulator and it received the highest score, better than the Sys-AgenT trained with the AgenT simulator, which is reasonable because through retrieval, Agenda-Retrieval (AgenR) simulators present more language diversity to the system during training. When interacting with a real user, the systems that can handle more language variations will do better.
The SL-based simulators received relatively low scores. Further  investigation on this cause is presented in the discussion section. 

``Auto-Success'' has been used to reflect the solved ratio previously. However, it's not necessarily correlated with the user-rated solved ratio. For example,  Sys-AgenG's Auto-Success rate is much higher than Sys-AgenR's Auto Success rate, but 
the users think that these two systems perform the same in terms of \textit{Solved Ratio}.

\noindent\textbf{Satisfaction.} Solving the user's problem doesn't necessarily lead to user satisfaction sometimes. It also depends on the system's efficiency and latency.  Therefore, besides Solved Ratio, we also directly ask Turkers how satisfied they are with the system. The result shows that among all systems, Sys-AgenR model received the highest score. The positive correlation between the ``Solved Ratio'' and ``Satisfaction'' in Table~\ref{table:simulator_results} also indicates automatic task completion rate is a good estimator for user satisfaction.

\noindent\textbf{Efficiency.} We directly ask Turkers how efficient the system is in solving their  problems since dialog length doesn't always correlate with the system efficiency. For example,  although the dialog length of Sys-AgenG and Sys-SLE are similar to each other,  users rated Sys-AgenG to be the most efficient one and Sys-SLE to be the most inefficient one. Again we suspect this is caused by different user communication pattern where some users prefer providing slots across multiple turns while others prefer providing all slots in one turn. 

\noindent\textbf{Naturalness.} We ask the Turkers to rate the naturalness of the system responses. All the systems share the same template-based NLG module designed by human experts, thus there shouldn't be a significant difference in the naturalness score. However, according to Table~\ref{table: human evaluation on AMT}, we find that the naturalness score seems to correlate with the overall system performance. A possible reason is that  the end-user is rating the system's naturalness by the overall performance instead of the system responses alone. When the dialog policy is bad, even if the NLG module can generate natural system responses, the users would still think the system is unnatural. 
This suggests that when designing dialog systems, NLG and policy selection modules should go hand-in-hand  in evaluation.

\noindent\textbf{Rule-likeness} We also ask the users to what extend they think the system is designed by handcrafted rules on a scale from 1 to 5, five means it is heavily handcrafted. Among all the models, Sys-SLT that is trained with the SL-Template simulator receives the lowest score, meaning it's the least rigid system. This is because SL-Template's dialog manager is learned with supervised learning, less rigid than the agenda-based dialog policy, which further leads to a less rigid behaviour of the trained dialog system.  

\subsection{Cross Study of Simulators and Systems}
From the last column in Table~\ref{table: human evaluation on AMT}, we find that although the automatic success rates claimed by the user simulator used to train the system are all relatively high, the high automatic success rate doesn't transfer to real human satisfaction.
In our setting, each simulator can be viewed as a new user with different communicating habits; therefore, we are curious to see if the performance can transfer to a different simulator when we test the RL system trained with simulator A against simulator B.
Table~\ref{tab:cross_test} shows a cross study between the six user simulators and the six systems trained with different simulators, where we fix the systems, have each simulator interact with each system for 200 episodes, and calculate the average success rate. The diagonal should reflect the ``Auto Success'' column in Table~\ref{table: human evaluation on AMT}, but since the 200 episodes are random and the ``Auto Success'' is the convergent success rate, the exact number won't be the same.

The last row in Table~\ref{tab:cross_test} shows the average success rate of each system across user simulators. There are some interesting findings.  1) Sys-AgenR that is trained with the Agenda-Retrieval simulator has the best average success rate, which agrees with the human evaluation on MTurk. 
2) A common practice to compare RL systems $S_1$, ... $S_n$ is to fix one user simulator $U$ and then compare the success rate of $S_1$, ..., $S_n$ on $U$. However, by looking at the fifth row for the SL-Retrieval simulator, it will prefer Sys-SLT (0.975) over Sys-AgenG (0.965), but actually the average performance of Sys-AgenG (0.882) is better than Sys-SLT (0.616) from the last row. 
This suggests that  when we want to compare two systems but don't have the resource to do human evaluation on the system performance, instead of solely comparing their success rates tested on  one simulator, we should build different types of user simulators and test the systems against multiple simulators to get a more holistic view of the systems. 
3) The diagonal in the table is usually the highest, meaning that RL policy does a good job optimizing for its own simulator but may not generalize to other user simulators. For example, the upper right corner performs the worst because the systems trained with SL-based simulators are worse in general, whose reason we will discuss later.

\subsection{Human Correlation Study}
\begin{figure}
\centering
  \includegraphics[width=0.98\columnwidth]{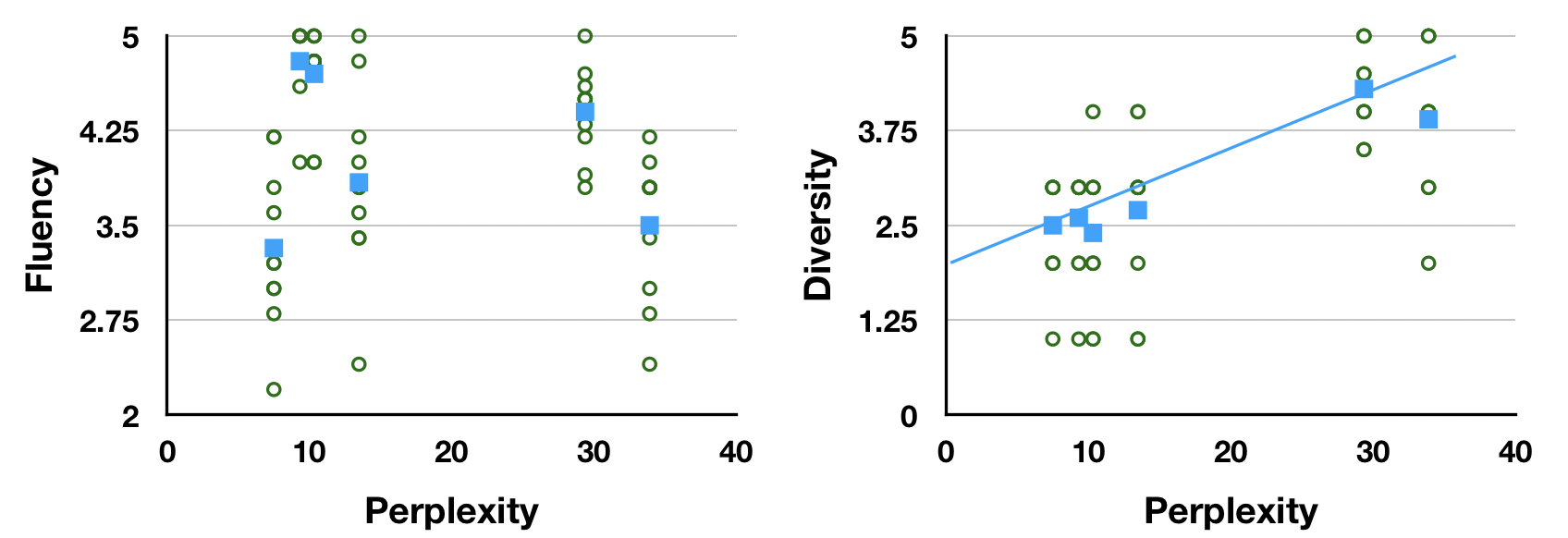}
  \caption{Correlation between sentence fluency and perplexity, and correlation between sentence diversity and perplexity. Green circles represent the human rated score, while the blue squares are the average score over different raters.}
  \label{fig:correlation}
\end{figure}

To test if the automatic metrics can reflect human evaluation, we compute the correlation between perplexity (PPL) and human evaluated fluency (Hu.Fl) and the correlation between perplexity and human evaluated diversity score (Hu.Div), which are $-0.21$ with $p>0.05$ and $0.95$ with $p=0.003$ respectively. We also visualize these metrics in Fig.~\ref{fig:correlation}. It shows that as an automatic metric, perplexity is a good estimator for language diversity but not for language fluency.


\section{Discussion and Future Work}
\begin{figure}
\centering
  \includegraphics[width=0.98\columnwidth]{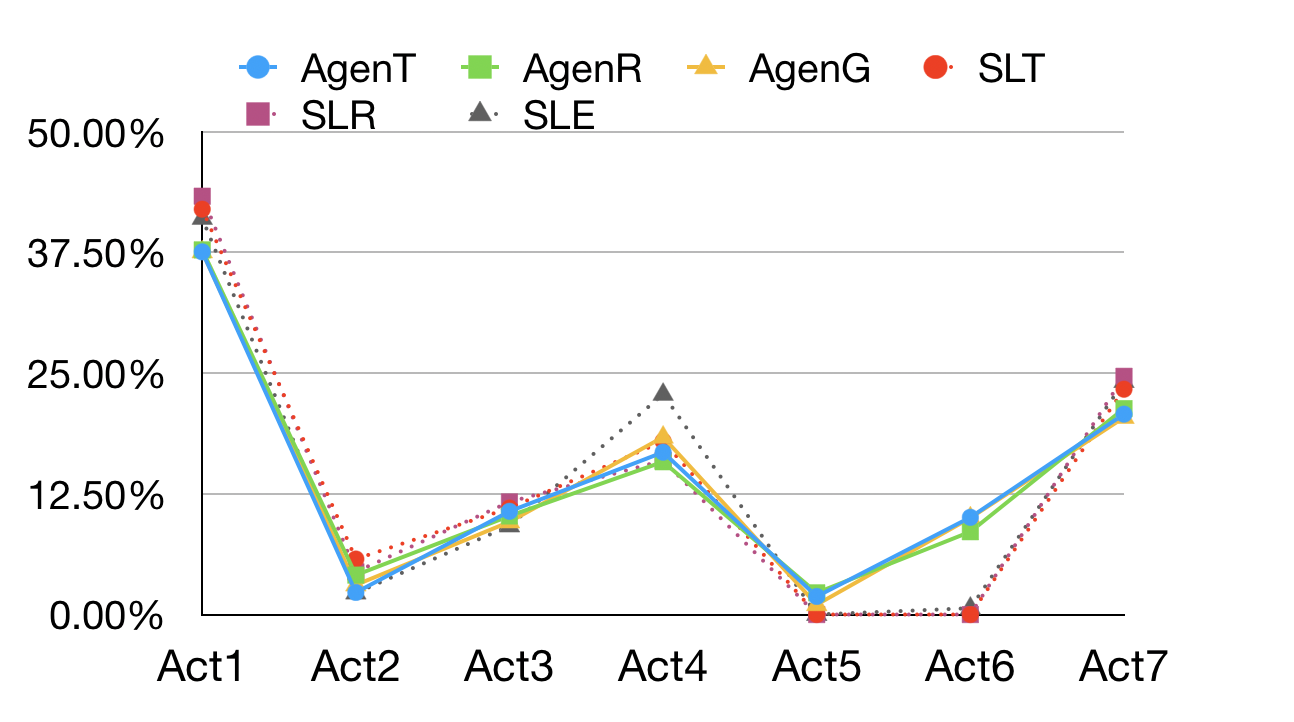}
  \caption{Dialog act distribution comparison. Act1 to Act7 corresponds to  the seven user dialog acts, \textit{``inform restaurant type''}, \textit{``inform restaurant type change''}, \textit{``ask info''}, \textit{``make reservation''}, \textit{``make reservation change time''}, \textit{``anything else''}, and \textit{``goodbye''}}
  \label{fig:act_distribution}
\end{figure}

SL-based simulators perform relatively worse than Agenda-based simulators when interacting with real users. We investigate the data and find it's caused by SL-based simulators  not exploring all possible paths. We draw the different dialog act distributions on simulated conversations in Fig.~\ref{fig:act_distribution}.  
For example, in agenda-based simulators, we explicitly have a rule for the dialog act ``anything else'' (Act6 in Fig.~\ref{fig:act_distribution}) but no such rules exist in SL-based simulators. Therefore, the RL model will experience the ``anything else'' scenario more in Agenda-based simulators than in SL-based simulators. When real users ask about ``anything else'', RL systems trained with Agenda-based simulators will have more experiences in handling such a case, compared to systems trained with SL-based simulators.



In this paper, we perform in-depth studies on the restaurant  domain as it's the most well-studied domain in task-oriented dialog systems, yet there's still no standard user simulator available. In the future we plan to include more domain using various domain-adaptive methods \cite{qian2019domain, tran2018adversarial, gavsic2017dialogue}  to support multi-domain dialog system research, and incorporate our work into more and more standardized dialog system platforms \cite{lee2019convlab}.

\section{Conclusions}

User simulators are essential components in training RL-based dialog systems. However, building user simulators is not a trivial task. In this paper, we surveyed through different ways to build user simulators at the levels of dialog manager and NLG, and analyzed the pros and cons of each method. Further, we evaluated each simulator with automatic metrics and human evaluations both directly and indirectly, and shared insights on better user simulator building based on comprehensive analysis.



\bibliography{emnlp-ijcnlp-2019}
\bibliographystyle{acl_natbib}


\appendix

\onecolumn
\section{Appendices}

\subsection{Generated Dialog Example}

\begin{figure*}[hbt!]
\centering
  \includegraphics[width=\textwidth, height=200px]{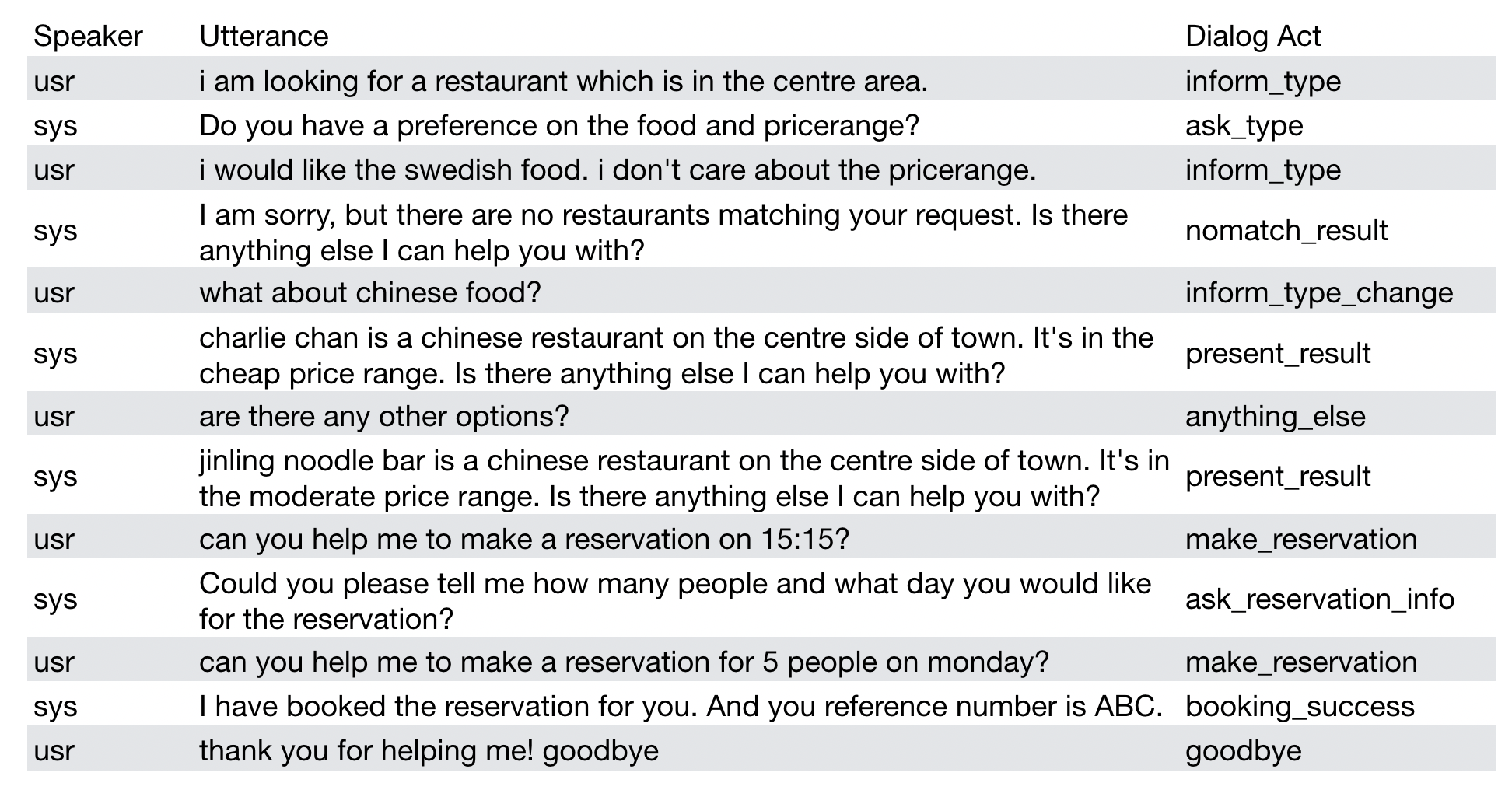}
  \caption{Generated dialog example}
\end{figure*}

\subsection{Human Evaluation Interface}

\begin{figure*}[hbt!]
\centering
  \includegraphics[width=\textwidth]{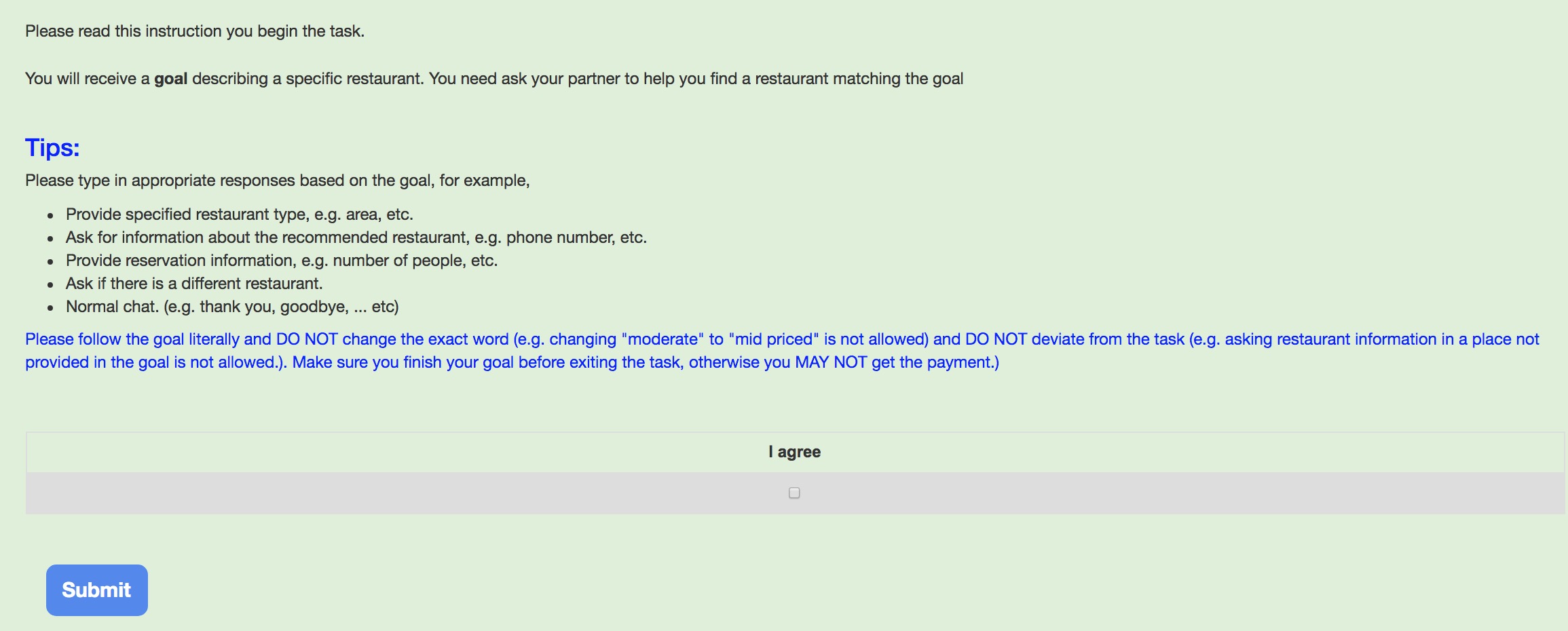}
  \caption{Task instructions for human evaluation}
  \label{fig:task_info}
\end{figure*}

\begin{figure*}[ht]
\centering
  \includegraphics[width=0.9\textwidth]{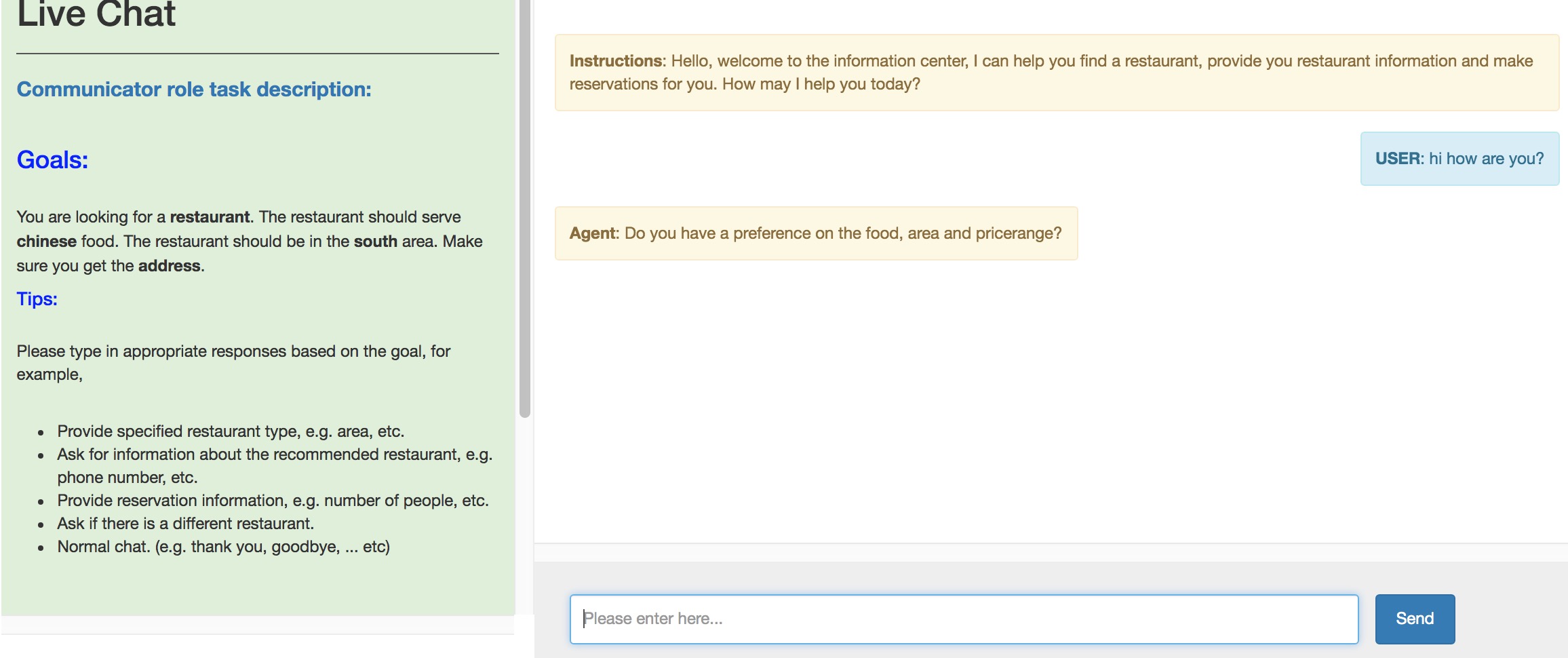}
  \caption{User chat interface}
  \label{fig:chat}
\end{figure*}

\begin{figure*}[ht]
\centering
  \includegraphics[width=0.9\textwidth]{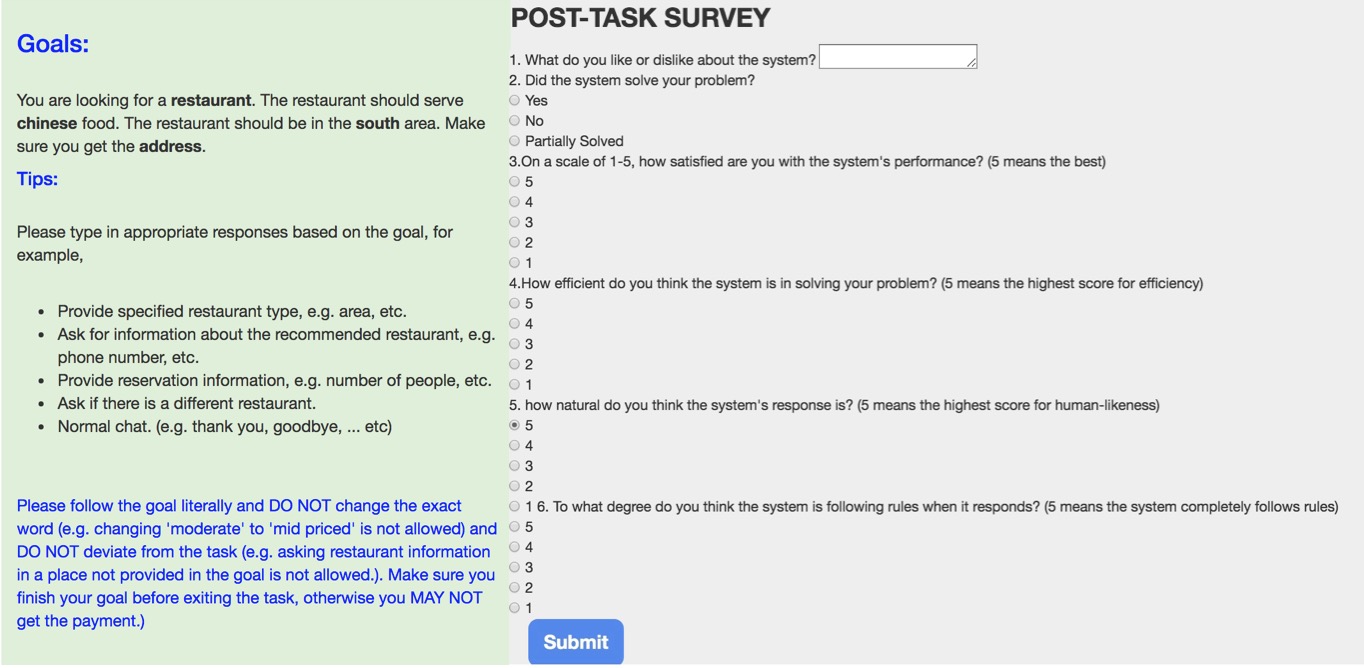}
  \caption{User survey interface}
  \label{fig:chat}
\end{figure*}

\end{document}